\documentclass{Interspeech}

\interspeechcameraready 

\title{``KAN you hear me?'' \\ Exploring Kolmogorov-Arnold Networks for Spoken Language Understanding}

\author[affiliation={1}]{Alkis}{Koudounas}
\author[affiliation={2}]{Moreno}{La Quatra}
\author[affiliation={1}]{Eliana}{Pastor}
\author[affiliation={3}]{Sabato Marco}{Siniscalchi}
\author[affiliation={1}]{Elena}{Baralis}

\affiliation{Politecnico di Torino}{Turin}{Italy}
\affiliation{Kore University of Enna}{Enna}{Italy}
\affiliation{Università degli Studi di Palermo}{Palermo}{Italy}
\email{
    alkis.koudounas@polito.it, moreno.laquatra@unikore.it
    }

\newcommand{\COne}{\textsc{FFF}}
\newcommand{\CTwo}{\textsc{KAN}}
\newcommand{\CThree}{\textsc{FFK}}
\newcommand{\CFour}{\textsc{FKK}}
\newcommand{\CFive}{\textsc{FKF}}
\newcommand{\CSix}{\textsc{FK}}

\usepackage{comment}
\usepackage{subfigure}
\usepackage{booktabs}
\usepackage{multirow}
\usepackage[table,xcdraw]{xcolor}
\usepackage[normalem]{ulem}
\usepackage{cite}
\useunder{\uline}{\ul}{}

\begin{document}

\maketitle

\begin{abstract}
Kolmogorov-Arnold Networks (KANs) have recently emerged as a promising alternative to traditional neural architectures, yet their application to speech processing remains under explored. 
This work presents the first investigation of KANs for Spoken Language Understanding (SLU) tasks. 
We experiment with 2D-CNN models on two datasets, integrating KAN layers in five different configurations within the dense block. 
The best-performing setup, which places a KAN layer between two linear layers, is directly applied to transformer-based models and evaluated on five SLU datasets with increasing complexity. 
Our results show that KAN layers can effectively replace the linear layers, achieving comparable or superior performance in most cases. 
Finally, we provide insights into how KAN and linear layers on top of transformers differently attend to input regions of the raw waveforms.
\end{abstract}

\section{Introduction}
Kolmogorov-Arnold Networks (KANs) have recently emerged as an alternative to traditional neural architectures, offering advantages in modeling complex, nonlinear relationships through learnable activation functions~\cite{liu2025kan}. 
Inspired by the Kolmogorov-Arnold representation theorem~\cite{kolmogorov1961representation, kolmogorov1957representation, braun2009constructive}, KANs replace fixed activation functions with learnable univariate functions, which enables them to approximate multivariate continuous functions. Preliminary works have demonstrated KAN's both advantages and limitations across various domains~\cite{yu2024kan, poeta2024benchmarking}. However, recent studies suggest that they can offer superior modeling capabilites in specific applications, including computer vision~\cite{azam2024suitability, cambrin2024kan}, medical image segmentation~\cite{li2024u}, and time-series forecasting~\cite{xu2024kolmogorov, han2024kan4tsf}. 

Despite their success in these areas, the application of KANs to speech-processing tasks remains largely under-explored. 
Spoken Language Understanding (SLU), in particular, is a fundamental component of human-computer interaction, enabling systems to extract meaning from spoken input and interpret user intentions~\cite{tur2011spoken, koudounas2024contrastive}. SLU plays a critical role in applications such as voice assistants, automated customer service, and voice-controlled smart devices, where accurate interpretation of natural language is essential for seamless interaction~\cite{towardse2eslu, prioritizingslu, sluedge}.
Although traditional neural networks have been widely adopted for SLU, the potential of KANs in this domain has yet to be investigated. 
We aim to bridge this gap by assessing whether the unique properties of KANs can be used to enhance performance on SLU tasks.

Recent studies have begun to explore the use of KANs in speech-related applications. For instance, \cite{xu2024effective} applied KANs to keyword spotting, highlighting their effectiveness in modeling high-level features in lower-dimensional space, while~\cite{li2024investigation} introduced KANs for speech enhancement tasks, demonstrating their ability to improve speech quality in both time and frequency domains. However, these studies have focused primarily on low-level acoustic tasks, leaving the application of KANs to higher-level semantic tasks, such as SLU, unexamined.

In this work, we present the first investigation of KANs for SLU tasks. We consider the following exploration dimensions.

\begin{itemize}
    \item 
\textbf{Integration options.}
We experiment with five different configurations, integrating KAN layers within the final dense block of a 2D-CNN-based model operating on spectrogram inputs. These evaluations are carried out on two datasets: \textsc{Fluent Speech Commands} (\textsc{FSC})~\cite{fsc} and \textsc{Timers and Such}~\cite{timersandsuch}. 
We also compare several function approximations within the KAN layer, finding that B-spline functions outperform Radial Basis Functions (RBF)~\cite{rbf, rbfkan}, Reflectional SWitch Activation Function (RSWAF)~\cite{rswaf}, Chebyshev~\cite{chebyshev}, and Group-Rational KAN (GR-KAN)~\cite{yang2024kolmogorov} alternatives.

\item \textbf{Task complexity.}
We apply the best-performing setup to two transformer-based models, wav2vec 2.0~\cite{wav2vec2} and XLS-R~\cite{babu2021xls}, and evaluate their performance across five datasets of increasing complexity: \textsc{FSC}, \textsc{Timers and Such}, \textsc{SLURP}~\cite{slurp}, \textsc{ITALIC}~\cite{italic} in Italian, and \textsc{SpeechMASSIVE}~\cite{speechmassive} in German and French.
Our experiments demonstrate that KAN layers can replace linear layers in SLU models, achieving comparable or superior performance in most cases without increasing model size or training time. 

\item \textbf{Error analysis.} Transformers equipped with KAN and linear layers differently attend to input regions of the raw waveform. We investigate cases where introducing a KAN layer corrects predictions, revealing patterns of behavior aligned with human reasoning.
\end{itemize}

Through this work, we aim to inspire further research into alternative neural architectures for speech processing, particularly in SLU applications. 

\begin{figure}[ht!]
    \centering
    \includegraphics[width=0.47\textwidth]{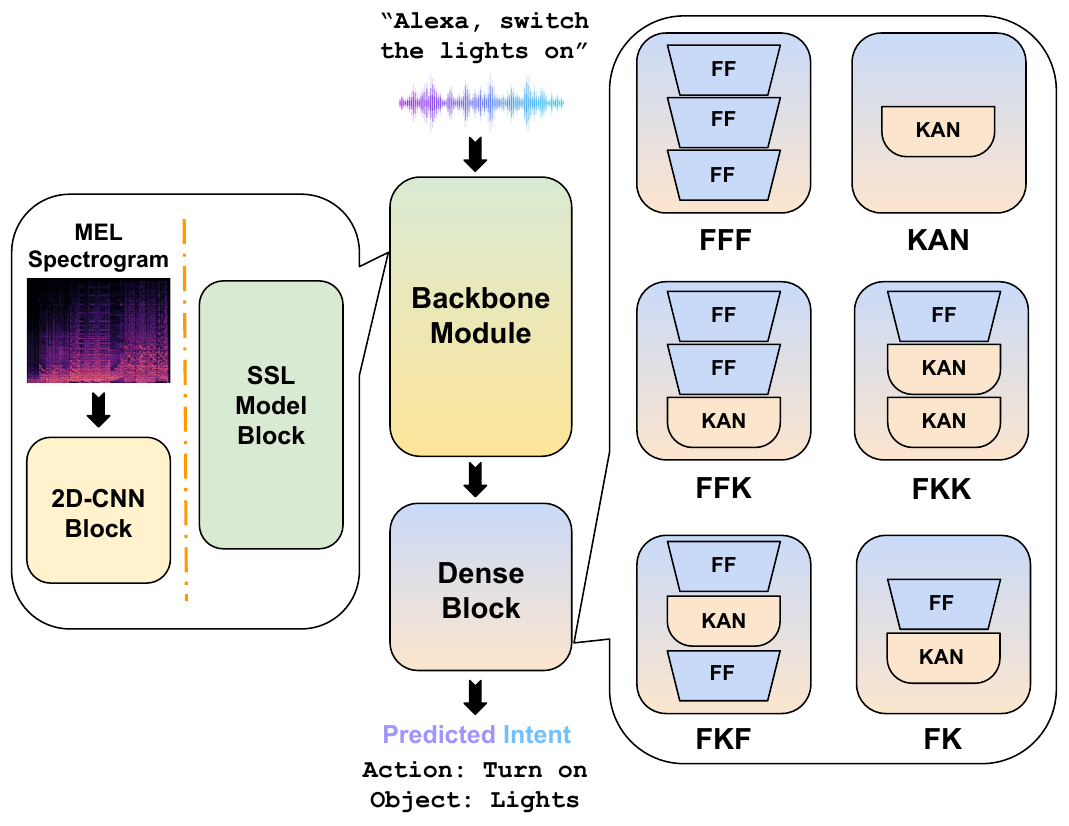}
    \caption{\textbf{Proposed configurations' overview.} \COne\ is the baseline with a fully-connected MLP. The other configurations show five alternatives differently integrating linear and KAN layers.}
    \label{fig:kan_architecture}
\end{figure}

\section{Preliminaries}
A KAN layer $L(x)$ models continuous functions by decomposing them into learnable univariate transformations. 
Formally, for an input vector $x \in \mathbb{R}^I$ and output dimension $O$, it can be expressed as:

\begin{equation}
    L(x) = \Phi \circ x = \prod_{i=1}^{I} \phi_{i,1}(x_i) \cdots \prod_{i=1}^{I} \phi_{i,O}(x_i)
\end{equation}

\noindent
where $\Phi$ is a matrix of learnable scalar functions $\phi_{i,o}(\cdot)$, with $i$ indexing input dimensions and $o$ indexing output dimensions.
For efficiency, a KAN layer is typically approximated using a combination of a basis function $(x)$ (e.g., Swish) and a \textit{B-spline} function, which enables smooth curve fitting with lower-degree polynomials:

\begin{equation}
    L(x) = w_1 b(x) + w_2 \text{spline}(x)
\end{equation}

\noindent where $w_1$ and $w_2$ are learnable scalars. This formulation enhances flexibility while maintaining computational efficiency. 
Several alternatives have been proposed to better approximate functions in KAN layers. 
RBF~\cite{rbf, rbfkan} provides smooth approximations by weighting basis functions centered at different points based on their distance from the input. 
Chebyshev polynomials~\cite{chebyshev} are a sequence of orthogonal polynomials defined recursively and commonly used in approximation theory due to their minimization of the maximum error, making them effective for function interpolation. 
RSWAF~\cite{rswaf} approximates B-splines using a modified version of the Switch Activation Function having Reflectional symmetry. 
Group-Rational KAN (GR-KAN)~\cite{yang2024kolmogorov} replaces B-splines with rational functions, aiming to enhance the expressiveness of the approximation through learnable rational transformations.

\section{KAN in SLU}
We first investigate the impact of architectural modifications on the dense classification block of a 2D-CNN, and then extend our analysis to transformer-based models.  

The baseline architecture (\COne\ in Fig.~\ref{fig:kan_architecture}) employs a 2D-CNN feature extractor followed by a dense block comprising a 3-layer Multi-Layer Perceptron (MLP). 
Each Feed-Forward (FF) layer within the MLP incorporates GELU activation, dropout regularization, and batch normalization. In formulas:
\begin{equation*}
    y = \rho(W_3 \cdot \text{Dropout}(\rho(W_2 \cdot \text{Dropout}(\rho(W_1 \cdot x + b_1)) + b_2)) + b_3)
\end{equation*}
where $x \in \mathcal{X}$ represents the input vector from the CNN feature extractor, $W_i \in \mathbb{R}^{n_i \times n_{i-1}}$  and $b_i \in \mathbb{R}^{n_i}$ are the weights and biases of the $i$-th layer, respectively, $\rho(\cdot)$ denotes the GELU activation function, and $y \in \mathcal{Y}$ is the output. 
This architecture serves as the basis for comparison with five configurations (Fig.~\ref{fig:kan_architecture}) designed to explore the interplay between linear (FF) and KAN layers.
We hypothesize that KAN layers can learn more efficient representations than linear layers by capturing nonlinear patterns directly. 
Through different configurations of FF and KAN layers, we aim to identify which structures best leverage their complementary strengths.

\vspace{1mm}
\noindent \textbf{\CTwo: KAN-Only.} This configuration replaces the entire 3-FF-layer MLP with a single KAN layer. While keeping the model parameters constrained, this simplification evaluates the raw representational capacity of a KAN layer. By directly mapping CNN-extracted features to the output space, we remove intermediate linear transformations, isolating the KAN layer’s ability to capture complex, non-linear relationships in the data.  

\vspace{1mm}
\noindent \textbf{\CThree\ and \CFour: Varying KAN Placement.} In configuration \CThree, we position a single KAN layer at the end of the block, preceded by two FF layers. In contrast, \CFour\ has one FF layer followed by two KAN layers. These designs investigate whether the KAN layer benefits from operating on a feature representation pre-processed by linear transformations. The FF layers might learn to disentangle or project the input into a space more tractable to the KAN-specific basis functions. 

\vspace{1mm}
\noindent \textbf{\CFive: Embedded KAN Layer.} Configuration \CFive\ embeds a single KAN layer within two FF layers. We hypothesize that this arrangement can offer a balance between the global representation learning capabilities of FF layers and the localized, non-linear feature extraction of the KAN layer. The initial FF layer might perform a coarse-grained feature extraction, preparing the input for the KAN layer to focus on specific, higher-order relationships. The final FF layer can then integrate these non-linear features with the original linear projections, potentially leading to a richer and more discriminative representation.

\vspace{1mm}
\noindent \textbf{\CSix: Minimal Hybrid Configuration.} Finally, \CSix\ represents the most compact hybrid architecture, comprising a single FF layer followed by a single KAN layer. This configuration serves as a baseline for evaluating the effectiveness of combining even just one FF and one KAN layer. It allows us to determine if the benefits of incorporating a KAN layer can be realized even with a minimal increase in architectural complexity. 

\vspace{1mm}
A key challenge in the original KAN implementation~\cite{liu2025kan} arises from the computational demands of expanding intermediate variables. 
For a layer with input dimension $d_{in}$ and output dimension $d_{out}$, the original approach requires expanding the input to a tensor of shape $(B, d_{out}, d_{in})$, to apply the activation functions, where $B$ is the batch size.
This expansion incurs significant memory overhead. 
However, recognizing that all activation functions within KAN are linear combinations of a fixed set of basis functions (e.g., B-splines), we exploit the more efficient recomputation available here\footnote{\url{https://github.com/Blealtan/efficient-kan}}. 
Instead of expanding the input, we activate it with different basis functions and subsequently combine the results linearly. 
This drastically reduces memory consumption and transforms the computation into a straightforward matrix multiplication, naturally compatible with both forward and backward passes.

We evaluate different function approximations within KAN layers, comparing B-splines against RBF~\cite{rbfkan}, Chebyshev~\cite{chebyshev} and RSWAF~\cite{rswaf} alternatives. 
After identifying the best configurations from our initial experiments on CNNs, we extend these setups to transformer-based models.  
This allows us to evaluate their generalization capabilities across different architectures. By applying the same architectural modifications, we assess whether the observed improvements hold beyond CNNs. 

\begin{table}[]
\centering
\caption{\textbf{2D-CNN Performance}. Performance comparison of several configurations with FF and KAN layers (as shown in Fig.~\ref{fig:kan_architecture}) on two SLU datasets. The baseline with a fully-connected MLP is highlighted in \colorbox[HTML]{DAE8FC}{light-blue}. Best results in \textbf{bold}, second-best \underline{underlined}. }
\vspace{-2mm}
\label{table-cnn2d}
\setlength{\tabcolsep}{4pt}
\scalebox{0.78}
{%
\begin{tabular}{@{}ccccccc@{}}
\toprule
    &       
    &                      
    & \multicolumn{2}{c}{\textbf{FSC}}          
    & \multicolumn{2}{c}{\textbf{Timers and Such}} \\ 
    \cmidrule(l){4-7} 

\multirow{-2}{*}{\textbf{Config}} &
  \multirow{-2}{*}{\textbf{Size}} &
  \multirow{-2}{*}{\begin{tabular}[c]{@{}c@{}}\textbf{Training} \\ \textbf{Time}\end{tabular}} &
  \multirow{1}{*}{\textbf{Accuracy}} &
  \multirow{1}{*}{\textbf{F1 Macro}} &
  \multirow{1}{*}{\textbf{Accuracy}} &
  \multirow{1}{*}{\textbf{F1 Macro}} \\ 
  \midrule
  
\rowcolor[HTML]{DAE8FC}  
\COne 
    & 6.2M  
    & 1.82it/s 
    & .555\scriptsize±.020 
    & .549\scriptsize±.022 
    & .471\scriptsize±.032 
    & .455\scriptsize±.042 \\

\CTwo 
    & 17.6M 
    & 1.71it/s
    & .452\scriptsize±.007          
    & .445\scriptsize±.008          
    & .416\scriptsize±.010            
    & .364\scriptsize±.019           \\

\CThree 
    & 6.2M  
    & 1.96it/s             
    & {\ul .559\scriptsize±.012}    
    & {\ul .550\scriptsize±.009}   
    & {\ul .565\scriptsize±.017}      
    & {\ul .533\scriptsize±.025}     \\

\CFour 
    & 6.4M  
    & 1.75it/s             
    & .533\scriptsize±.009          
    & .527\scriptsize±.010          
    & .553\scriptsize±.007            
    & .515\scriptsize±.009           \\

\CFive 
    & 6.3M  
    & 1.95it/s            
    & \textbf{.583\scriptsize±.006}
    & \textbf{.575\scriptsize±.002} 
    & \textbf{.571\scriptsize±.011}   
    & \textbf{.545\scriptsize±.022}  \\

\CSix
    & 6.2M  
    & 1.86it/s 
    & .526\scriptsize±.015          
    & .521\scriptsize±.013          
    & .533\scriptsize±.029            
    & .497\scriptsize±.032 \\ 
    \bottomrule
\end{tabular}}
\end{table}

 \begin{figure}[ht!]
    \centering
    \subfigure[Fixed hidden size.] {\includegraphics[width=0.21\textwidth]{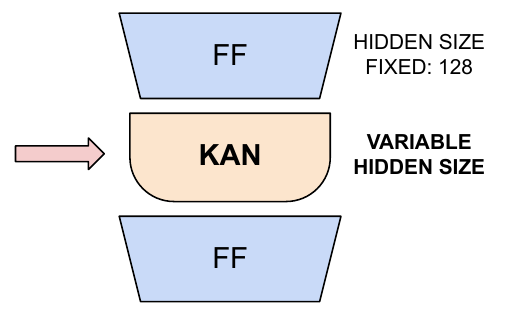}
    \label{fig-ablation-hs-a}} 
    \hspace{0pt}
    \subfigure[Variable hidden size.] {\includegraphics[width=0.21\textwidth]
    {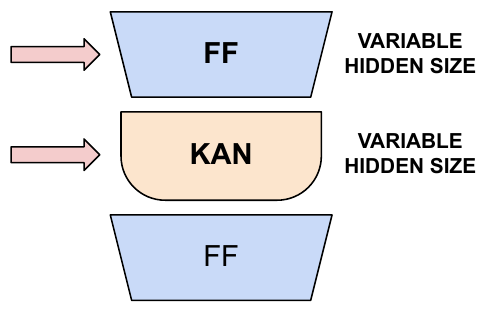}
    \label{fig-ablation-hs-b}} 
    \caption{\textbf{Ablation on hidden size.} Fixed (a) and variable (b) hidden size study on \CFive\ configuration.}
    \label{fig-ablation-hs}
\end{figure}

\section{Experimental Setup}

\vspace{1mm}
\noindent \textbf{Datasets.} 
We evaluate our approach on five publicly available intent classification datasets: \textsc{FSC}~\cite{fsc}, Timers and Such~\cite{timersandsuch}, and SLURP~\cite{slurp} for English, \textsc{ITALIC}~\cite{italic} for Italian, and \textsc{Speech-MASSIVE}~\cite{speechmassive} for German and French. The first two datasets are relatively simple, containing 31 and 4 intents, respectively. In contrast, \textsc{SLURP}, \textsc{ITALIC}, and \textsc{Speech-MASSIVE} are significantly larger, comprising 60 intents and offering greater linguistic diversity. \textsc{ITALIC} and \textsc{Speech-MASSIVE} can be considered multilingual extensions of \textsc{SLURP}, covering Italian, German, and French, respectively.  

Among these datasets, \textsc{Timers and Such} is unique in that it consists of spoken commands specifically designed for common voice control scenarios involving numerical inputs.  

\vspace{1mm}
\noindent \textbf{Models.} We first train a 2D-CNN using Mel spectrograms as input. The architecture consists of four convolutional layers with progressively increasing output channels (16, 32, 64, and 128). Each layer employs a kernel size of 3, a stride of 1, and a padding of 2. Following each convolution, we apply two-dimensional batch normalization, GeLU activation function, and dropout. Max pooling is applied after the second and fourth convolutional layers to downsample the feature maps.
A fully connected classification block follows the convolutional layers, comprising three linear layers, each followed by layer normalization, GeLU, and dropout. When replacing a feed-forward with a KAN layer, we substitute the entire block consisting of the linear layer, normalization, activation function, and dropout.
Mel spectrograms are computed using 400 FFT size, a window length of 400, a hop length of 160, and 64 Mel filters. The model is trained for a maximum of 20 epochs with an early stopping criterion of 5 epochs. We use a batch size of 256, an initial learning rate of 1e-4, and AdamW optimizer with weight decay. Learning rate adjustments are handled via a plateau scheduler based on validation accuracy.

We also evaluate transformer models, specifically the monolingual wav2vec 2.0\footnote{\texttt{\href{https://huggingface.co/facebook/wav2vec2-base}{huggingface.co/facebook/wav2vec2-base}}} and the multilingual XLS-R\footnote{\texttt{\href{https://huggingface.co/facebook/wav2vec2-xls-r-300m}{huggingface.co/facebook/wav2vec2-xls-r-300m}}}. These models utilize the same dense classification block as the 2D-CNN. Their training setup follows the same configuration as described before, with a reduced batch size of 64, 5e-5 learning rate, and a maximum of 50 epochs.

The code to reproduce our experiments is available at \texttt{\url{https://github.com/koudounasalkis/SLU-KAN}}.

\section{Results}

In the following, we present the wide variety of experiments performed to explore the behavior of KAN network integration.

\subsection{Experimental results on 2D-CNN} 
Table~\ref{table-cnn2d} shows the results on the 2D-CNN.
\COne\ serves as the baseline with only FF layers, providing a reference point for evaluating KAN-based configurations. 
\CTwo, which fully replaces FF layers with a single KAN layer, shows a notable drop in performance while increasing the number of parameters. This suggests that KAN alone struggles to stabilize feature transformations, and FF layers remain fundamental for structured representation learning. 
\CThree, where a KAN layer is placed at the end of the FF stack, slightly improves performance, indicating that linear feature extraction first, followed by a nonlinear transformation with KAN, is beneficial. 
\CFour, where 2 KAN layers follow a single FF layer, performs slightly worse than \CThree, suggesting that KAN provides a very effective intermediate layer, but is less effective as final layer.
\CFive, which places KAN between two FF layers, achieves the best accuracy and F1 scores across both datasets while also converging slightly faster, showing that a balance between structured transformations from FF layers and flexible representations from KAN is optimal for SLU tasks. 
\CSix, where KAN is placed after a single FF layer, leads to lower performance similar to \CFour. 
Overall, KAN layers provide improvements when properly integrated with FF layers, as seen in \CFive, while full replacement, as in \CTwo, proves ineffective.

\begin{table}[]
\centering
\caption{\textbf{Ablation on hidden size}. Performance comparison of \COne\ and \CFive\ configurations, the latter with fixed hidden size (as shown in Fig.\ref{fig-ablation-hs-a}) and variable hidden size (Fig.\ref{fig-ablation-hs-b}). The baseline is highlighted in \colorbox[HTML]{DAE8FC}{light-blue}.}
\vspace{-2mm}
\label{table-ablation-hs}
\setlength{\tabcolsep}{4pt}
\scalebox{0.90}
{%
\begin{tabular}{@{}cccccc@{}}
\toprule
\multicolumn{1}{c}{} 
    &      
    &      
    &          
    & \multicolumn{2}{c}{\textbf{FSC}} \\
    \cmidrule(l){5-6} 
\multicolumn{1}{c}{\multirow{-2}{*}{\textbf{Config}}} 
    & \multirow{-2}{*}{\begin{tabular}[c]{@{}c@{}}\textbf{Hidden}\\\textbf{Size}\end{tabular}} 
    & \multirow{-2}{*}{\begin{tabular}[c]{@{}c@{}}\textbf{Model}\\\textbf{Size}\end{tabular}} 
    & \multirow{-2}{*}{\begin{tabular}[c]{@{}c@{}}\textbf{Training}\\\textbf{Time}\end{tabular}} 
    & \multirow{1}{*}{\textbf{Accuracy}} 
    & \multirow{1}{*}{\textbf{F1 Macro}} \\
\midrule

\rowcolor[HTML]{DAE8FC} 

\COne
    & 128  
    & 6.2M 
    & 1.82it/s 
    & .555\scriptsize±.020          
    & .549\scriptsize±.022 \\

\CFive(a)  
    & 32   
    & 6.2M 
    & 1.97it/s 
    & .554\scriptsize±.022
    & .533\scriptsize±.025 \\
    
\CFive(a)  
    & 64 
    & 6.2M 
    & 1.97it/s 
    & .565\scriptsize±.011
    & .554\scriptsize±.012 \\
    
\CFive(a)  
    & 128 
    & 6.3M
    & 1.95it/s 
    & \textbf{.583\scriptsize±.006} 
    & \textbf{.575\scriptsize±.002} \\
    
\CFive(a)  
    & 256  
    & 6.5M 
    & 1.91it/s 
    & .578\scriptsize±.019
    & .562\scriptsize±.023 \\
    
\CFive(a)  
    & 512  
    & 6.9M 
    & 1.89it/s 
    & .552\scriptsize±.013
    & .544\scriptsize±.010 \\
    
\CFive(a)    
    & 1024 
    & 7.7M 
    & 1.85it/s 
    & .547\scriptsize±.010
    & .543\scriptsize±.009 \\ 
\midrule

\CFive(b)   
    & 32   
    & 1.6M  
    & 3.01it/s 
    & .513\scriptsize±.013          
    & .493\scriptsize±.014 \\
    
\CFive(b)   
    & 64   
    & 3.2M  
    & 2.69it/s 
    & .533\scriptsize±.007          
    & .520\scriptsize±.004 \\
    
\CFive(b)   
    & 128  
    & 6.3M  
    & 1.95it/s 
    & .583\scriptsize±.006          
    & .575\scriptsize±.002 \\
    
\CFive(b)   
    & 256  
    & 12.9M 
    & 1.80it/s 
    & .586\scriptsize±.013          
    & .584\scriptsize±.017 \\
    
\CFive(b)   
    & 512  
    & 27.4M 
    & 1.71it/s 
    & \textbf{.631\scriptsize±.014} 
    & \textbf{.628\scriptsize±.016} \\
    
\CFive(b)   
    & 1024 
    & 61.2M 
    & 1.52it/s 
    & .616\scriptsize±.029          
    & .579\scriptsize±.032 \\
    
\bottomrule
\end{tabular}}
\end{table}

\vspace{2mm}
\noindent \textbf{Analysis of hidden size.} We analyze how the hidden size impacts the final performance in Table~\ref{table-ablation-hs}. \CFive(a) maintains a fixed hidden size of 128 for the initial FF layer while varying the KAN layer's output size. The results show that performance improves as the KAN output size increases up to 128, where accuracy and F1-score peak. 
Beyond this, performance shows a slight decline, indicating that very large KAN outputs might affect feature learning stability. 
Training time slightly decreases with smaller KAN output sizes but remains stable across configurations. 
\CFive(b), where both FF and KAN hidden sizes are variable, exhibits a different trend. With small hidden sizes (32 and 64), performance is significantly lower, showing that a minimal feature space limits the network’s ability to model complex speech patterns. As the hidden size increases, accuracy and F1-score improve, surpassing \CFive(a) at 512 hidden units, where it achieves the highest scores. However, performance slightly drops at 1024, potentially due to overfitting or inefficient representation scaling. Model size and training time increase notably in \CFive(b), with the largest configuration reaching over 60M params, suggesting that while increasing model capacity enhances performance, it comes with higher computational costs.

\begin{table}[]
\centering
\caption{\textbf{Ablation on approximation function}. Comparison of \COne\ and \CFive\ configurations, the latter with different approximation functions. The baseline is highlighted in \colorbox[HTML]{DAE8FC}{light-blue}.}
\vspace{-2mm}
\label{table-ablation-function}
\setlength{\tabcolsep}{4pt}
\scalebox{0.90}
{%

\begin{tabular}{@{}cccccc@{}}
\toprule
\multicolumn{1}{c}{} 
    & 
    & 
    & 
    & \multicolumn{2}{c}{\textbf{FSC}} \\
    \cmidrule(l){5-6}
\multicolumn{1}{c}{\multirow{-2}{*}{\textbf{Config}}} 
    & \multirow{-2}{*}{\begin{tabular}[c]{@{}c@{}}\textbf{Approx.}\\\textbf{Function}\end{tabular}} 
    & \multirow{-2}{*}{\begin{tabular}[c]{@{}c@{}}\textbf{Model}\\\textbf{Size}\end{tabular}} 
    & \multirow{-2}{*}{\begin{tabular}[c]{@{}c@{}}\textbf{Training}\\\textbf{Time}\end{tabular}}
    & \multirow{1}{*}{\textbf{Accuracy}}
    & \multirow{1}{*}{\textbf{F1 Macro}} \\
\midrule
\rowcolor[HTML]{DAE8FC} 
\COne 
    & - 
    & 6.2M 
    & 1.82it/s 
    & .555\scriptsize±.020 
    & .549\scriptsize±.022 \\
    
\CFive 
    & B-Spline
    & 6.3M 
    & 1.95it/s 
    & \textbf{.583\scriptsize±.006} 
    & \textbf{.575\scriptsize±.002} \\

\CFive   
    & RBF                           
    & 6.3M 
    & 1.98it/s 
    & .554\scriptsize±.005          
    & .540\scriptsize±.007 \\
    
\CFive & 
    RSWAF                         
    & 6.3M 
    & 1.98it/s 
    & .538\scriptsize±.009          
    & .526\scriptsize±.011 \\

\CFive 
    & Chebyshev
    & 6.2M 
    & 1.65it/s 
    & .569\scriptsize±.010 
    & .564\scriptsize±.009 \\

\CFive 
    & GR-KAN
    & 6.2M 
    & 1.65it/s 
    & .563\scriptsize±.009 
    & .554\scriptsize±.011 \\
    
\bottomrule
\end{tabular}}
\vspace{-1mm}
\end{table}

\begin{table}[]
\centering
\caption{\textbf{Transformer models performance}. Comparison in terms of F1 macro of \COne, \CThree, and \CFive\ configurations 
on five SLU datasets. Monolingual wav2vec 2.0 model for the first block of datasets, multilingual XLS-R for the second. The baseline is highlighted in \colorbox[HTML]{DAE8FC}{light-blue}.}
\vspace{-2mm}
\label{table-transformers}
\setlength{\tabcolsep}{4pt}
\scalebox{0.95}
{%
\begin{tabular}{ccccc}
\toprule

\textbf{Config}
    & \textbf{Size}
    & \textbf{FSC}
    & \textbf{Timers and Such}
    & \textbf{SLURP} \\
    \midrule

\rowcolor[HTML]{DAE8FC} 
\COne &
  94.6M &
  .990\scriptsize±.002 &
  .976\scriptsize±.006 &
  \textbf{.539\scriptsize±.010} \\
\CThree &
  94.7M &
  .994\scriptsize±.001 &
  .985\scriptsize±.003 &
  .534\scriptsize±.006 \\
\CFive &
  95.4M &
  \textbf{.995\scriptsize±.001} &
  \textbf{.992\scriptsize±.004} &
  .531\scriptsize±.001 \\
  \midrule\midrule

\multirow{2.5}{*}{\textbf{Config}} 
    & \multirow{2.5}{*}{\textbf{Size}} 
    & \multirow{2.5}{*}{\textbf{ITALIC}} &
  \multicolumn{2}{c}{\textbf{SPEECH-MASSIVE}} \\
  \cmidrule(l){4-5} 

&
   & 
   & \textbf{de-DE}
   & \textbf{fr-FR}\\
   \midrule

\rowcolor[HTML]{DAE8FC} 
\COne &
  315.8M &
  .747\scriptsize±.004 &
  .659\scriptsize±.011 &
  .672\scriptsize±.008 \\
\CThree &
  315.9M &
  .734\scriptsize±.007 &
  .642\scriptsize±.012 &
  .678\scriptsize±.010 \\
\CFive &
  316.5M &
  \textbf{.749\scriptsize±.006} &
  \textbf{.667\scriptsize±.009} &
  .\textbf{679\scriptsize±.004} \\ 
  \bottomrule
  
\end{tabular}}
\vspace{-6mm}
\end{table}

\smallskip
\noindent \textbf{Analysis of approximation function.}
Table \ref{table-ablation-function} analyzes diverse approximation functions.
Among the different approximation functions tested in \CFive, B-Spline achieves the highest scores.  
RBF and RSWAF show lower performance, indicating that not all approximation functions contribute positively in this context.  
Chebyshev and GR-KAN perform better than RBF and RSWAF, but both remain below B-Spline.  
The choice of approximation function thus significantly impacts performance, with B-Spline resulting the most effective in this setup.

\begin{figure}[ht]
    \centering
    \includegraphics[width=0.47\textwidth]{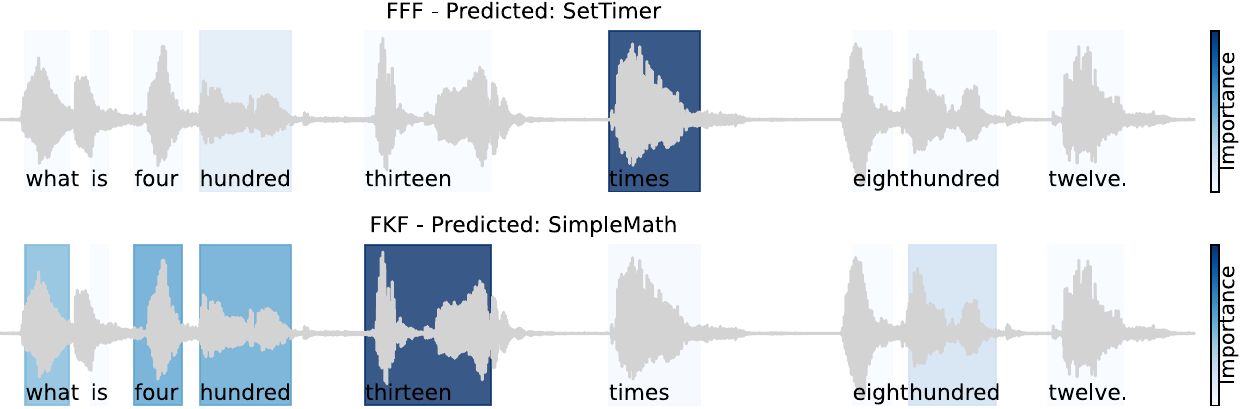}
    \caption{\textbf{Example of word-level 
    explanations} for \COne\ and \CFive\ predictions; \textsc{Timers and Such}, SimpleMath intent.}
    \label{fig:explanations}
    \vspace{-5mm}
\end{figure}

\subsection{Experimental results with transformers}
Table~\ref{table-transformers} presents a performance comparison of three configurations (baseline \COne, \CThree, and \CFive) using transformer-based models across five SLU datasets. 
The first block corresponds to monolingual wav2vec 2.0 models, while the second block includes multilingual \textsc{XLS-R} models. 
The results show that \CThree\ and \CFive\ consistently outperform \COne\ in most datasets, though the differences are often marginal. For \textsc{FSC} and \textsc{Timers and Such}, \CThree\ and \CFive\ reach near-perfect F1 scores. 
In \textsc{SLURP}, all models perform similarly, suggesting that the dataset may pose intrinsic challenges that are not addressed by architectural modifications. 
In the multilingual setting, performance gaps are slightly less pronounced.  
For \textsc{ITALIC} and \textsc{Speech-MASSIVE}, \CFive\ achieves the best results, suggesting better generalization across languages. Despite these improvements, training times are nearly identical across configurations, indicating that the modifications in \CThree\ and \CFive\ do not introduce significant computational overhead. 
The results demonstrate that the configuration identified for 2D-CNN models and directly applied to transformer-based models generalizes well across datasets. This suggests that the architectural refinements in \CFive\ are robust and transferable across different model types.

\vspace{-1mm}
\subsection{Attention to input regions}
\vspace{-1mm}
We investigate whether introducing a KAN layer affects how the model attends to input data. 
We focus on transformer models and \textsc{Timers and Such}, where the \CFive\ configuration shows a marked improvement in performance. 
We analyze differences in how \COne\ and \CFive\ models process inputs, particularly in cases where \COne\ makes incorrect predictions that \CFive\ corrects. 
To explore this, we use an explanation technique that assigns relevance scores to word-level segments, indicating how much each spoken word impacts the prediction~\cite{pastor-etal-2024-explaining}. 
Fig.~\ref{fig:explanations} provides an example: \COne\ incorrectly classifies an utterance to `SetTimer', with its explanation showing that the model attends to the word ``\textit{times}". 
In contrast, \CFive\ correctly classifies it to the `SimpleMath' intent, focusing on numerical cues. 
This pattern holds across other misclassified samples, where \COne\ consistently assigns relevance to ``\textit{times}", suggesting it as a source of ambiguity, while \CFive\ attends to more pertinent words. \CFive's explanations align more closely with human reasoning, making them plausible~\cite{jacovi-goldberg-2020-towards} and suggesting KAN layers can be a valid alternative to linear ones.

\vspace{-1mm}
\section{Conclusion}
\vspace{-1mm}
This work explores the integration of KAN layers in SLU tasks, demonstrating their effectiveness across different model architectures and languages. 
Our experiments show that strategic placement of KAN layers between FF layers achieves optimal performance while maintaining computational efficiency comparable to traditional approaches. 
Multilingual generalization and the plausibility of model explanation suggest promising directions for future research in semantic speech processing.

\section{Acknowledgments}
This work is partially supported by the FAIR - Future Artificial Intelligence Research (PIANO NAZIONALE DI RIPRESA E RESILIENZA (PNRR) – MISSIONE 4 COMPONENTE 2, INVESTIMENTO 1.3 – D.D. 1555 11/10/2022, PE00000013)  and the spoke ``FutureHPC \& BigData'' of the ICSC - Centro Nazionale di Ricerca in High-Performance Computing, Big Data and Quantum Computing, both funded by the European Union - NextGenerationEU, and by the "D.A.R.E. – Digital Lifelong Prevention" project (code: PNC0000002, CUP: B53C22006450001), co-funded by the Italian Complementary National Plan PNC-I.1 Research initiatives for innovative technologies and pathways in the health and welfare sector (D.D. 931 of 06/06/2022) and by the European Union – Next Generation EU under the National Recovery and Resilience Plan (PNRR) – M4 C2, Investment 1.1: Fondo per il Programma Nazionale di Ricerca e Progetti di Rilevante Interesse Nazionale (PRIN) - PRIN 2022 - "SHAPE-AD" (CUP: J53D23007240008).
This manuscript reflects only the authors' views and opinions, neither the European Union nor the European Commission can be considered responsible for them. 

\bibliographystyle{IEEEtran}
\bibliography{mybib}

\end{document}